\documentclass[11pt]{article}
\usepackage{emnlp2016}
\usepackage{times}

\emnlpfinalcopy 
\def\emnlppaperid{1067} 

\usepackage{url}
\usepackage{latexsym}
\usepackage{mathtools}
\usepackage{graphicx}
\usepackage{amsmath}
\usepackage{xspace}
\usepackage{subcaption}
\usepackage{tikz-dependency}
\usepackage{balance}
\usepackage{enumitem} 
\usepackage{pifont}
\usepackage{float}
\usepackage{multirow}
\usepackage{xspace}
\usepackage{soul}
\usepackage{censor}
\usepackage{xargs} 
\usepackage[colorinlistoftodos,prependcaption,textsize=tiny]{todonotes}
\usepackage{marginnote}

\makeatletter
\newcommand{\@BIBLABEL}{\@emptybiblabel}
\newcommand{\@emptybiblabel}[1]{}
\makeatother
\usepackage{hyperref}
\hypersetup{
    colorlinks,
    linkcolor={blue!25!black},
    citecolor={blue!25!black},
    urlcolor={blue!25!black}
}

\usepackage{cleveref}
\Crefformat{equation}{(#2#1#3)}

\usepackage{microtype}
\usepackage{booktabs}
\usepackage{rotating}
\newcommandx{\todoir}[2][1=]{\todo[inline]{SR: #2}\xspace}
\newcommandx{\todoib}[2][1=]{\todo[inline]{YB: #2}\xspace}
\newcommandx{\todoijh}[2][1=]{\todo[inline]{JH: #2}\xspace}
\newcommandx{\todoij}[2][1=]{\todo[inline]{JB: #2}\xspace}
\newcommandx{\todor}[2][1=]{\todo[linecolor=red,backgroundcolor=red!25,bordercolor=red,#1]{SR: #2}\xspace}
\newcommandx{\todob}[2][1=]{\todo[linecolor=cyan,backgroundcolor=cyan!25,bordercolor=cyan,#1]{YB: #2}\xspace}
\newcommandx{\todojh}[2][1=]{\todo[linecolor=blue,backgroundcolor=blue!10,bordercolor=blue,#1]{JH: #2}\xspace}
\newcommandx{\todoj}[2][1=]{\todo[linecolor=green,backgroundcolor=green!25,bordercolor=green,#1]{JB: #2}\xspace}

\newcommand \ignore[1]{}

\usepackage{ccg}

\newcommand \gp{\textsc{GraphParser}\xspace}
\newcommand \cw{\textsc{ClueWeb09}\xspace}

\newcommand \CORPUS{\textsc{SPADES}\xspace}

\title{Evaluating Induced CCG Parsers on Grounded Semantic Parsing}
\author{\bf Yonatan Bisk$^1$\thanks{\ \ Equal contribution}\ \ \ \  Siva Reddy$^2$\footnotemark[1] \ \ \ {John Blitzer$^3$} \ \ \ Julia Hockenmaier$^4$\ \ \ {Mark Steedman$^2$}\\
$^{1}$ISI, University of Southern California\\
$^{2}$ILCC, School of Informatics, University of Edinburgh\\
$^{3}$Google, Mountain View\\
$^{4}$Department of Computer Science, University of Illinois at Urbana-Champaign\\
\href{mailto:ybisk@isi.edu}{\nolinkurl{ybisk@isi.edu}},
\href{mailto:siva.reddy@ed.ac.uk}{\nolinkurl{siva.reddy@ed.ac.uk}},
\href{mailto:blitzer@google.com}{\nolinkurl{blitzer@google.com}}, \\
\href{mailto:juliahmr@illinois.edu}{\nolinkurl{juliahmr@illinois.edu}},
\href{mailto:steedman@inf.ed.ac.uk}{\nolinkurl{steedman@inf.ed.ac.uk}},
}
\date{}

\begin{document}

\maketitle
\begin{abstract}
We compare the effectiveness of four different syntactic CCG parsers for a semantic slot-filling task to explore how much syntactic supervision is required for downstream semantic analysis.  
This extrinsic, task-based evaluation provides a unique window to explore the strengths and weaknesses of semantics captured by unsupervised grammar induction systems.
We release a new Freebase semantic parsing dataset called SPADES (\textbf{S}emantic \textbf{PA}rsing of \textbf{DE}clarative \textbf{S}entences) containing 93K cloze-style questions paired with answers.
We evaluate all our models on this dataset.
Our code and data are available at \url{https://github.com/sivareddyg/graph-parser}.
\end{abstract}

\section{Introduction}
The past several years have seen significant progress in unsupervised grammar induction \cite{Carroll:1992uba,Yuret-1998,Klein:2004uo,Spitkovsky:2010tn,Garrette:2015wt,Bisk:2015:ACL}. But how useful are unsupervised syntactic parsers for downstream NLP tasks? What phenomena are they able to capture, and where would additional annotation be required? Instead of standard intrinsic evaluations -- attachment scores that depend strongly on the particular annotation styles of the gold treebank -- we examine the utility of unsupervised and weakly supervised parsers for semantics. We perform an  extrinsic evaluation of unsupervised and weakly supervised CCG parsers on a grounded semantic parsing task that will shed light on the extent to which these systems recover semantic information. We focus on English to perform a direct comparison with supervised parsers (although unsupervised or weakly supervised approaches are likely to be most beneficial for domains or languages where supervised parsers are not available).

\begin{figure*}
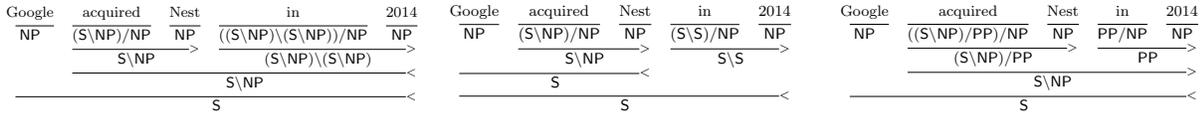

\scriptsize
\hspace{-1em}
\begin{subfigure}[b]{0.34\textwidth}
\resizebox{166pt}{!}{
\deriv{5}{
{\rm Google}&{\rm acquired}& {\rm Nest} & {\rm in} & \rm{2014} \\
\uline{1}& \uline{1}    & \uline{1} & \uline{1}    & \uline{1} \\
\cf{NP}   &\cf{(S\bs NP)/NP} &\cf{NP} & \cf{((S\bs NP)\bs (S\bs NP))/NP} & \cf{NP} \\
         &     \fapply{2} &  \fapply{2} \\
         & \mc{2}{\cf{S\bs NP}} & \mc{2}{\cf{(S\bs NP)\bs (S\bs NP)}} \\
         &     \bapply{4} \\
         & \mc{4}{\cf{S\bs NP}} \\
     \bapply{5} \\
     \mc{5}{\cf{S}} 
}
}
\caption{\scriptsize \textsl{\textbf{in 2014}} modifies \textsl{\textbf{acquired Nest}}}
\label{fig:multipleCCGa}
\end{subfigure}
~\hspace{-1em}
\begin{subfigure}[b]{0.34\textwidth}
\centering
\resizebox{140pt}{!}{
\deriv{5}{
{\rm Google}&{\rm acquired}& {\rm Nest} & {\rm in} & \rm{2014} \\
\uline{1}& \uline{1}    & \uline{1} & \uline{1}    & \uline{1} \\
\cf{NP}   &\cf{(S\bs NP)/NP} &\cf{NP} & \cf{(S\bs S)/NP} & \cf{NP} \\
         &     \fapply{2} &  \fapply{2} \\
         & \mc{2}{\cf{S\bs NP}} & \mc{2}{\cf{S\bs S}} \\
         \bapply{3} \\
         \mc{3}{\cf{S}} \\
      \bapply{5} \\
      \mc{5}{\cf{S}} 
}
}
\caption{\scriptsize \textsl{\textbf{in 2014}} modifies \textsl{\textbf{Google acquired Nest}}}
\label{fig:multipleCCGb}
\end{subfigure}
~\hspace{-1em}
\begin{subfigure}[b]{0.34\textwidth}
\resizebox{146pt}{!}{
\deriv{5}{
{\rm Google}&{\rm acquired}& {\rm Nest} & {\rm in} & \rm{2014} \\
\uline{1}& \uline{1}    & \uline{1} & \uline{1}    & \uline{1} \\
\cf{NP}   &\cf{((S\bs NP)/PP)/NP} &\cf{NP} & \cf{PP/NP} & \cf{NP} \\
         &     \fapply{2} &  \fapply{2} \\
	 & \mc{2}{\cf{(S\bs NP)/PP}} & \mc{2}{\cf{PP}} \\
         & \fapply{4} \\
         & \mc{4}{\cf{S\bs NP}} \\
      \bapply{5} \\
      \mc{5}{\cf{S}} 
}
}
\caption{\scriptsize \textsl{\textbf{acquired Google}} takes the argument \textsl{\textbf{in 2014}}}
\label{fig:multipleCCGc}
\end{subfigure}
\caption{Example of multiple valid derivations that can be grounded to the same Freebase logical form (Eq. \ref{eq:logical}) even though they differ dramatically in performance under parsing metrics (5, 4, or 3 ``correct" supertags).}
\label{fig:multipleCCG}
\end{figure*}


Specifically, we evaluate different parsing scenarios with varying amounts of supervision. These are designed to shed light on the question of how well syntactic knowledge correlates with performance on a semantic evaluation. 
We evaluate the following scenarios (all of which assume POS-tagged input): 1)~no supervision; 2)~a lexicon containing words mapped to CCG categories; 3)~a lexicon containing POS tags mapped to CCG categories;  4)~sentences annotated with CCG derivations (i.e., fully supervised).
Our evaluation reveals which constructions are problematic for unsupervised parsers (and annotation efforts should focus on). 
Our results indicate that unsupervised syntax is useful for semantics, while a simple semi-supervised parser outperforms a fully unsupervised approach, and could hence be a viable option for low resource languages.

\section{CCG Intrinsic Evaluations}
CCG~\cite{Steedman:2000tt} is  a  lexicalized  formalism in which words are assigned syntactic
types, also known as \textit{supertags}, encoding  subcategorization  information. Consider the
sentence \textsl{Google acquired Nest in 2014}, and its CCG derivations shown in
\Cref{fig:multipleCCG}.
In (a) and (b), the supertag of \textsl{acquired}, \cf{(S\bs NP) / NP},
indicates that it has two arguments, and the prepositional phrase \textsl{in 2014} is an adjunct,
whereas in (c) the supertag \cf{((S\bs NP)/PP)/NP} indicates \textsl{acquired} has three arguments
including the prepositional phrase. In (a) and (b), depending on the supertag of \textsl{in}, the
derivation differs. When trained on labeled treebanks, (a) is 
preferred. However note that all these derivations could lead to the same semantics
(e.g., to the logical form in Equation~\ref{eq:logical}).
Without syntactic supervision, there may not be any reason for the parser to prefer one analysis over the other. 
One procedure to evaluate unsupervised induction methods has been to compare the
assigned supertags to treebanked supertags, but this evaluation does not consider that multiple derivations could lead to the same semantics.
This problem is also not solved by evaluating syntactic dependencies. 
Moreover, while many dependency standards agree on the head 
direction of simple constituents (e.g., noun phrases) they disagree on the most
semantically useful ones (e.g., coordination and relative clauses).\footnote{Please see \newcite{Bisk:2013vu} for more details.}  


\section{Our Proposed Evaluation}
The above syntax-based evaluation metrics conceal the real performance
differences and their effect on downstream tasks. Here we propose an extrinsic evaluation where we evaluate our ability to convert sentences to Freebase logical forms starting via CCG derivations. Our motivation is that most sentences can only have a single realization in Freebase, and any derivation that could lead to this realization is potentially a correct derivation. For example, the Freebase logical form for the example sentence in \Cref{fig:multipleCCG} is shown below, and none of its derivations are penalized if they could result in this logical form.


\setlength{\abovedisplayskip}{-30pt}
\setlength{\belowdisplayskip}{0pt}
\begin{equation}
\begin{small}
\begin{aligned}
  & \lambda e.\, \mathrm{business.acquisition}(e) \\
  & \wedge \mathrm{acquiring\_company}(e, \textsc{Google}) \\
  & \wedge \mathrm{company\_acquired}(e, \textsc{Nest}) \\
  & \wedge \mathrm{date}(e, \textsc{2014})
\end{aligned}
\end{small}
\label{eq:logical}
\end{equation}


\begin{figure*}
\centering
\begin{tabular}{@{}c@{\hspace{1em}}c@{}}
\begin{minipage}[c]{0.50\textwidth}
\resizebox{\textwidth}{!}{
\deriv{5}{
  {\rm Google}&{\rm acquired}& \langle{\tt blank}\rangle & {\rm which} & \rm{was\, founded\, in\, PA} \\
\uline{1}& \uline{1} & \uline{1} & \uline{1} & \uline{1} \\
\cf{NP} & \cf{(S\bs NP)/NP} & \cf{NP} & \cf{ (NP \bs NP)/ (S\bs NP)} & \cf{S\bs NP} \\
         &  &   &  \fapply{2} \\
	 & & & \mc{2}{\cf{NP\bs NP}} \\
         & & \bapply{3} \\
         & & \mc{3}{\cf{NP}} \\
	 & \fapply{4} \\
	 & \mc{4}{\cf{S\bs NP}} \\
        \bapply{5} \\
        \mc{5}{\cf{S}} 
}
}
\end{minipage} & 
\begin{minipage}[c]{0.3\textwidth} 
\includegraphics[width=\linewidth]{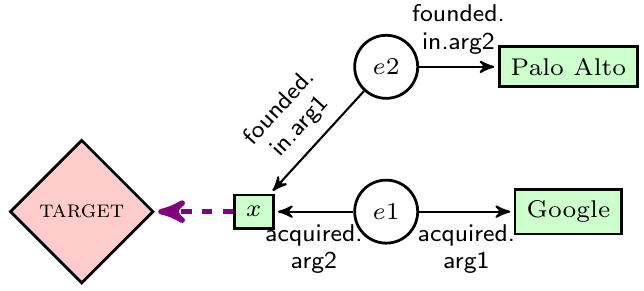}
\end{minipage}
\\
\multicolumn{2}{c}{
\scriptsize $\lambda e_1. \exists x e_2.\, \textsc{target}(x) \wedge \mathrm{acquired}(e_1) \wedge arg_1(e_1, \mathrm{Google}) \wedge arg_2(e_1, x)$  $\wedge\, \mathrm{founded}(e_2) \wedge \mathbf{arg_2(e_2}, x) \wedge \mathrm{in}(e_2, \mathrm{PaloAlto})$} \\[5px]

\begin{minipage}[c]{0.50\textwidth}
\resizebox{\textwidth}{!}{
\deriv{5}{
  {\rm Google}&{\rm acquired}& \langle {\tt blank}\rangle & {\rm which} & \rm{was\, founded\, in\, PA} \\
\uline{1}& \uline{1} & \uline{1} & \uline{1} & \uline{1} \\
\cf{NP} & \cf{(S\bs NP)/NP} & \cf{NP} & \cf{ ((S \bs NP)\bs(S \bs NP))/ (S\bs NP)} & \cf{S\bs NP} \\
         & \fapply{2}   &  \fapply{2} \\
	 & \mc{2}{\cf{S\bs NP}} & \mc{2}{\cf{(S \bs NP)\bs S\bs NP}} \\
         & \bapply{4} \\
         & \mc{4}{\cf{S\bs NP}} \\
        \bapply{5} \\
        \mc{5}{\cf{S}} 
    }
}
\end{minipage} &
\begin{minipage}[c]{0.3\textwidth} 
\includegraphics[width=\linewidth]{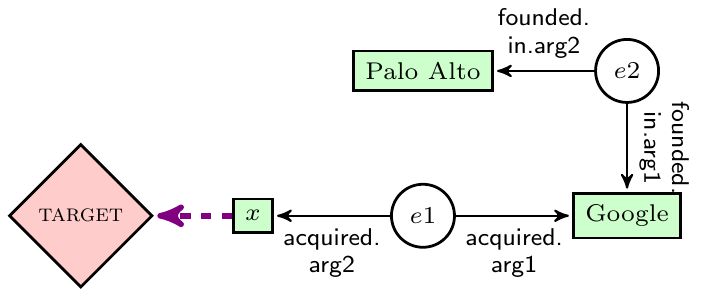}
\end{minipage} \\
\multicolumn{2}{c}{\scriptsize{$\lambda e_1. \exists x e_2.\, \textsc{target}(x) \wedge \mathrm{acquired}(e_1) \wedge arg_1(e_1, \mathrm{Google}) \wedge arg_2(e_1, x) \wedge\, \mathrm{founded}(e_2) \wedge \mathbf{arg_2(e_2}, \mathrm{Google}) \wedge \mathrm{in}(e_2, \mathrm{PaloAlto})$}}
\end{tabular}
  \caption{The lexical categories for \textsl{which} determine the relative clause attachment and therefore the resulting ungrounded logical form.  The top derivation correctly executes a query to retrieve companies founded in Palo Alto and acquired by Google.  The bottom incorrectly asserts that Google was founded in Palo Alto.
  }
  \label{fig:groundingExample}
\end{figure*}


Since grammar induction systems are traditionally trained on declarative sentences, we would ideally require declarative sentences paired with Freebase logical forms. But such datasets do not exist in the Freebase semantic parsing literature \cite{cai-yates:2013:*SEM,berant-EtAl:2013:EMNLP}. To alleviate this problem, and yet perform Freebase semantic parsing, we propose an entity slot-filling task.

\paragraph{Entity Slot-Filling Task.} Given a declarative sentence containing mentions of Freebase entities, we randomly remove one of the mentions to create a blank slot.  The task is to fill this slot by translating the declarative sentence into a Freebase query.
Consider the following sentence where the entity \textsl{Nest} has been removed:

\sethlcolor{black}
\begin{center}
\begin{small}
\textsl{Google acquired \underline{\hspace{3em}} which was founded in Palo Alto}
\end{small}
\end{center}
To correctly fill in the blank, one has to query Freebase for the entities acquired by \textsl{Google} (constraint~1) and founded in \textit{Palo Alto} (constraint~2). If either of those constraints are not applied, there will be many entities as answers. For each question, we execute a single Freebase query containing all the constraints and retrieve a list of answer entities. From this list, we pick the first entity as our predicted answer, and consider the prediction as correct if the gold answer is the same as the predicted answer. 

\section{Sentences to Freebase Logical Forms}
CCG provides a clean interface between syntax and semantics, i.e. each argument of a word’s syntactic category corresponds to an argument of the lambda expression that defines its semantic interpretation  (e.g., the lambda expression corresponding to the category \cf{(S\bs NP)/NP} of the verb \textsl{acquired} is $\lambda f. \lambda g. \lambda e. \exists x. \exists y. \mathrm{acquired}(e) \wedge f(x) \wedge g(y) \wedge arg_1(e,y) \wedge arg_2(e,x)$), and the logical form for the complete sentence can be constructed by composing word level lambda expressions following the syntactic derivation \cite{bos2004wide}. In \Cref{fig:groundingExample} we show two syntactic derivations for the same 
sentence, and the corresponding logical forms and equivalent graph representations derived by \gp \cite{Reddy:2014wc}. The graph representations are possible because \gp assumes access to co-indexations of input CCG categories. We provide co-indexation for all induced categories, including multiple co-indexations when an induced category is ambiguous. For example, \cf{(S\bs N)/ (S\bs N)} refers to either \cf{(S_{x}\bs N_{y})/ (S_{x}\bs N_{y})} indicating an auxiliary verb or \cf{(S_{x}\bs N_{y})/ (S_{z}\bs N_{y})} indicating a control verb.
Initially, the predicates in the expression/graph will be based entirely on the surface form of the words in the sentence.  
This is the ``\textit{ungrounded}'' semantic representation. 

Our next step is to convert these ungrounded graphs to Freebase graphs.\footnote{Note that there is one-to-one correspondence between Freebase graphs and Freebase logical forms.} Like \newcite{Reddy:2014wc}, we treat this problem as a graph matching problem. Using \gp we retrieve all the Freebase graphs that are isomorphic to the ungrounded graph, and select only the graphs that could correctly predict the blank slot, as candidate graphs. Using these candidate graphs, we train a structured perceptron that learns to rank grounded graphs for a given ungrounded graph.\footnote{Please see Section~4.3 of \newcite{reddy_transforming_2016} for details.} We use ungrounded predicate and Freebase predicate alignments as our features. 

\section{Experiments}
\subsection{Training and Evaluation Datasets}
Our dataset \CORPUS (\textbf{S}emantic \textbf{PA}rsing of \textbf{DE}clarative \textbf{S}entences) is constructed from the declarative sentences collected by \newcite{Reddy:2014wc} from \cw \cite{FACC1} based on the following constraints: 
1)~There exists at least one isomorphic Freebase graph to the ungrounded representation of the input sentence; 2)~There are no variable nodes in the ungrounded graph (e.g., \textsl{Google acquired a company} is discarded whereas \textsl{Google acquired the company Nest} is selected). We split this data into training (85\%), development (5\%) and testing (10\%) sentences (Table \ref{tab:corpus_stats}). We introduce empty slots into these sentences by randomly removing an entity.  
\CORPUS can be downloaded at \url{http://github.com/sivareddyg/graph-parser}.  

\begin{table}
\centering
\begin{small}
\begin{tabular}{lrrrr}
\toprule
      & Sentences & Tokens & Types & Entities \\
\midrule
Train & 79,247 & 685,922 & 69,095 & 37,606 \\
Dev   & 4,763  &  41,102 & 9,306  & 4,358\\
Test  & 9,309  &  80,437 & 15,180 & 7,431\\
\bottomrule
\end{tabular}
\end{small}
\caption{\CORPUS Corpus Statistics}
\label{tab:corpus_stats}
\end{table}

\begin{table*}[htp]
  \centering
  \begin{small}
  \begin{tabular}{l@{\hspace{4pt}}l@{\hspace{2em}}cc@{\hspace{2em}}cccc}
  \toprule
  	& 						& \multicolumn{2}{@{\hspace{-1em}}l@{\hspace{3em}}}{CCGbank (Syntax)} & \multicolumn{4}{c}{Slot Filling (Semantics)} \\
    &                         & LF1 & UF1  & 2 & 3 & 4 & Overall \\ 
    & Sentences 			  &     &      & $\sim$6K & $\sim$3K & $\sim$600 & $\sim$10K \\
    \midrule
    & Bag-of-Words			& --   & --   & 34.1 & 28.1 & \bf 16.5 & 31.4 \\
    \midrule
    \multirow{4}{*}{\begin{turn}{90}Syntax\end{turn}} & \textsc{Unsupervised}            & 37.1 & 64.2 &  27.7 & 20.0 &\bf 16.8  & 24.8  \\
    & \textsc{Semi-Supervised-PoS}  & 53.0 & 68.5 & 30.5  & 21.8  &\bf 18.8 & 27.3 \\
    & \textsc{Semi-Supervised-Word} & 53.5 & 68.9 & 30.7  & 25.5  & \bf       18.7 &  28.4  \\
    & \textsc{Supervised}           & 84.2 & 91.0 & 32.7 & 29.0 &\bf 20.2 & 30.9 \\
    \bottomrule
  \end{tabular}
  \end{small}
  \caption{Syntactic and semantic evaluation of the parsing models. Left: Simplified labeled F1 and undirected unlabeled F1 on CCGbank, Section 23. Right: Slot filling performance (by number of entities per sentence). {\color{red} Slot-filling results are updated after the camera-ready submission. In the previous version, instead of evaluating if the gold entity is same as the first predicted entity, we mistakenly evaluated if the gold entity is present in the list of predicted answer entities. However, the initial claims are still valid. All other results and discussion are revised.}} 
  \label{tab:testEval}
  \vspace{0.5em}
\end{table*}

There has been other recent interest in similar datasets for sentence completion \cite{zweig-EtAl:2012:ACL2012} and machine reading \cite{NIPS2015_5945}, but unlike other corpora our data is tied directly to Freebase and requires the execution of a semantic parse to correctly predict the missing entity.  This is made more explicit by the fact that one third of the entities in our test set are never seen during training, so without a general approach to query creation and execution there is a limit on a system's performance.

\subsection{Our Models}
We use different CCG parsers varying in the amounts of supervision. For the \textsc{Unsupervised} scenario, we use \newcite{Bisk:2015:ACL}'s parser which exploits a small set of universal rules to automatically induce and weight a large set of lexical categories. For the semi-supervised, we explore two options -- \textsc{Semi-Supervised-Word} and \textsc{Semi-Supervised-PoS}. We use Bisk et al.~in both settings but we constrain its lexicon manually rather than inducing it from scratch. In the former, we restrict the top 200 words in English to occur only with the CCG categories that comprise 95\% of the occurrences of a word's use in Section 22 of WSJ/CCGbank. In the latter, we restrict the POS tags instead of words. For the \textsc{Supervised} scenario, we use EasyCCG \cite{lewis-steedman:2014:EMNLP2014} trained on CCGbank. 

Finally, in order to further demonstrate the amount of useful information being learned by our parsers, 
we present a competitive Bag-of-Words baseline, which is a perceptron classifier that performs ``semantic parsing" by predicting either a Freebase or a null relation between the empty slot and every other entity in the sentence, using the words in the sentence as features.
This naive approach is competitive on simple sentences with only two entities, rivaling even the fully supervised parser, but falters as complexity increases.

\subsection{Results and Discussion}
Our primary focus is a comparison of intrinsic syntactic evaluation with our 
extrinsic semantic evaluation.  To highlight the differences we present 
Section 23 parsing performance for our four models in Table~\ref{tab:testEval}.  
Dependency performance is evaluated on both the simplified labeled F1 of 
\newcite{Bisk:2015:ACL} and Undirected Unlabeled F1.

Despite the supervised parser performing almost twice as well as the semi-supervised parsers on CCGbank LF1 (53.5 vs 84.2), in our semantic evaluation we see a comparatively small gain in performance (28.4 vs 30.9).  It is interesting that such weakly 
supervised models are able to achieve over 90\% of the performance of a
fully supervised parser.  To explore this further, we
break down the semantics performance of all our models 
by the number of entities in a 
sentence.  Each sentence has two, three, or four entities, one of which will be dropped for 
prediction.  The more entities there are in a sentence, the more likely the models are to 
misanalyze a relation leading to their making the wrong prediction.  These results are presented on
the right side of Table \ref{tab:testEval}. There are still notable discrepancies in performance, 
which we analyze more closely in the next section.

Another interesting result is the drop in performance by the Bag-of-Words Model.  As the number of entities in the sentence increase, the model weakens, performing worse than the unsupervised parser on sentences with four entities.  It becomes non-trivial for it to isolate which entities and relations should be used for prediction.  
This seems to indicate that the unsupervised grammar is capturing more useful syntactic/semantic information than what is available from the words alone. 
Ensemble systems that incorporate syntax and a Bag-of-Words baseline may yield even better performance.

\subsection{The Benefits of Annotation}

\begin{figure}[t]
\centering
\includegraphics[width=0.9\linewidth]{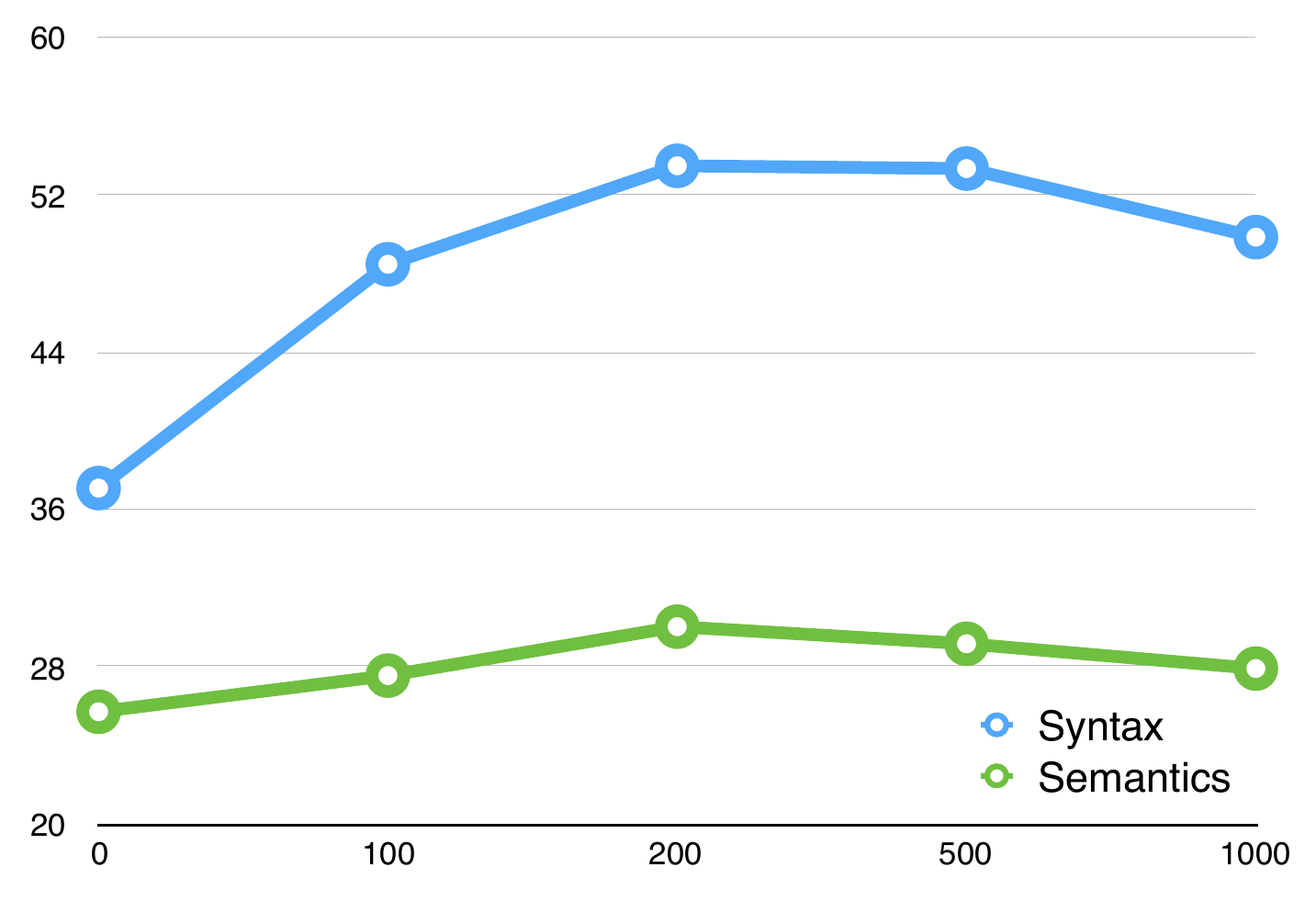}
\caption{The effect of increasing number of lexical types in \textsc{Semi-Supervised-Word} on syntax and semantics. When the lexicon grows past 200 lexical types both syntax and semantics begins to degrade. We also observe there is a correlation between syntactic parsing and semantic parsing performance.}
\label{fig:annotation}
\end{figure}

The performance of \textsc{Semi-Supervised-PoS} and \textsc{Semi-Supervised-Word} suggests that when resources are scarce, it is beneficial to create a even a small lexicon of CCG categories.
We analyze this further in Figure \ref{fig:annotation}.  Here we show how performance changes as a function of the number of labeled lexical types.  Our values range from 0 to 1000 lexical types.  We see syntactic improvements of 16pts and semantic gains of 4.33pts (16.8\%) with 200 words, before performance degrades.  It is possible that increasing annotation may only benefit fully supervised models. Finally, when computing the most frequent lexical types we excluded commas.  We found a drop in performance when restricting commas to the category \cf{,} (they are commonly \cf{conj} in our data).  Additional in-domain knowledge might further improve performance.


\begin{table}[t]
\centering
\begin{small}
\begin{tabular}{@{}l@{\hspace{5pt}}l@{\hspace{5pt}}l@{}}
\toprule
 & Error  & Example \\
\midrule
\multirow{4}{*}{\begin{turn}{90}Prevalent\end{turn}} & Incorrect conjunction & \textit{Stockholm, Sweden} \\
& Appositive & \textit{, a chemist ,} \\
& Introductory clauses & \textit{In Frankfurt, ...} \\
& Reduced relatives & \textit{... , established in 1909, ...} \\
\midrule
\multirow{3}{*}{\begin{turn}{90}B\&H 15\end{turn}}& Verb chains & \textit{is also headquartered} \\
& Possessive & \textit{Anderson 's Foundation} \\
& PP Attachment & \textit{of the foundation in Vancouver}\\
\bottomrule
\end{tabular}
\end{small}
\caption{Causes of semantic grounding errors with examples not previously isolated via intrinsic evaluation.}
\label{tab:analysis}
\end{table}

\subsection{Common Errors}

\newcite{Bisk:2015:ACL} performed an in-depth analysis of the types of categories learned and correctly used by their models (the same models as this paper).  Their analysis was based on syntactic evaluation against CCGbank.  In particular, they found the most egregious ``semantic'' errors to be the misuse of verb chains, possessives and PP attachment (bottom of Table~\ref{tab:analysis}).  Since we now have access to a purely semantic evaluation, we can therefore ask whether these errors exist here, and how common they are.  We do this analysis in two steps.  First,  we manually analyzed parses for which the unsupervised model failed to predict the correct semantics, but where the supervised parser succeeded.  The top of Table~\ref{tab:analysis} presents several of the most common reasons for failure.  These mistakes were more mundane (e.g. incorrect use of a conjunction) than failures to use complex CCG categories or analyze attachments.


Second, we can compare grammatical decisions made by the semi-supervised and unsupervised parsers against EasyCCG on sentences they successfully grounded. 
\newcite{Bisk:2015:ACL} found that their unsupervised parser made mistakes on many very simple categories.  We found the same result.  When evaluating our parsers against the treebank we found the unsupervised model only correctly predicted transitive verbs 20\% of the time and adverbs 39\% of the time.  In contrast, on our data, we produced the correct transitive category (according to EasyCCG) 65\% of the time, and the correct adverb 68\% of the time.  These correct parsing decisions also lead to improved performance across many other categories (e.g. prepositions).
This is likely due to our corpus containing simpler constructions. In contrast, auxiliary verbs, relative clauses, and commas still proved difficult or harder than in the treebank.  This implies that future work should tailor the annotation effort to their specific domain rather than relying on guidance solely from the treebank.

\section{Conclusion}

Our goal in this paper was to present the first semantic evaluation of induced grammars in order to better understand their utility and strengths.  
We showed that induced grammars are learning more semantically useful structure  than a Bag-of-Words model. Furthermore, we showed how minimal syntactic supervision can provide substantial gains in semantic evaluation. Our ongoing work explores creating a syntax-semantics loop where each benefits the other with no human (annotation) in the loop.

\section*{Acknowledgments}
This paper is partly based on work that was done when the first and second authors were interns at Google, and on work that that was supported by NSF grant 1053856 to JH, and a Google PhD Fellowship to SR.

\bibliography{bibliography}
\bibliographystyle{emnlp2016}

\end{document}